\documentclass[journal]{IEEEtran}
\pdfoutput=1

\usepackage{multirow}

\usepackage{cite}

\ifCLASSINFOpdf
   \usepackage[pdftex]{graphicx}
   \graphicspath{{../pdf/}{../jpeg/}}
\else
\fi
\usepackage{amsmath}
\ifCLASSOPTIONcompsoc
  \usepackage[caption=false,font=normalsize,labelfont=sf,textfont=sf]{subfig}
\else
  \usepackage[caption=false,font=footnotesize]{subfig}
\fi
\begin{document}
%
\title{Channel Compression: Rethinking Information Redundancy among Channels in CNN Architecture}
%
%
%

\author{Jinhua Liang,~\IEEEmembership{Student Member,~IEEE,},
		Tao~Zhang*,~\IEEEmembership{Member,~IEEE,},
        and~Guoqing~Feng
}

%
%

\markboth{}%
{}
%



\maketitle

\begin{abstract}
Model compression and acceleration are attracting increasing attentions due to the demand for embedded devices and mobile applications. Research on efficient convolutional neural networks (CNNs) aims at removing feature redundancy by decomposing or optimizing the convolutional calculation. In this work, feature redundancy is assumed to exist among channels in CNN architectures, which provides some leeway to boost calculation efficiency. Aiming at channel compression, a novel convolutional construction named compact convolution is proposed to embrace the progress in spatial convolution, channel grouping and pooling operation. Specifically, the depth-wise separable convolution and the point-wise interchannel operation are utilized to efficiently extract features. Different from the existing channel compression method which usually introduces considerable learnable weights, the proposed compact convolution can reduce feature redundancy with no extra parameters. With the point-wise interchannel operation, compact convolutions implicitly squeeze the channel dimension of feature maps. To explore the rules on reducing channel redundancy in neural networks, the comparison is made among different point-wise interchannel operations. Moreover, compact convolutions are extended to tackle with multiple tasks, such as acoustic scene classification, sound event detection and image classification. The extensive experiments demonstrate that our compact convolution not only exhibits high effectiveness in several multimedia tasks, but also can be efficiently implemented by benefiting from parallel computation.
\end{abstract}


%
\IEEEpeerreviewmaketitle

\section{Introduction}
%
%
%
%
\IEEEPARstart{C}{onvolutional} neural networks (CNNs) are attracting considerable attention in an increasing array of area, such as computer vision\cite{8371638,8669681,8735820}, computational acoustics\cite{8922774,ZHANG2020113067,7952132} and natural language processing\cite{8734068,Ma_2015_ICCV,CHEN2017221}. The general trend is to design deeper and more complicated network architecture to pursue better performance. However, massive resources are required for desired performance, which limits CNN-based classifiers from the real-time inference in mobile applications. Over the past few decades, various methods have been exploited for model compression and acceleration, including pruning\cite{LiPruning,HeChannel,Liu2017Learning,YeRethinking}, weight sharing\cite{Chen2015Compressing,sironi2014learning}, low-rank matrix factorization\cite{sainath2013low,jaderberg2014speeding,denil2013predicting} and knowledge distillation\cite{hinton2015distilling,luo2016face,balan2015bayesian}.  Despite their desirable compression ability, most of the compression methods typically suffer from two major drawbacks. First, the original complex model is replaced with an approximation one, resulting in the error accumulation. Therefore, fine-tuning is usually necessary for satisfying performance. Second, various manually chosen parameters (even a lot of empirical engineering that only experts competent to deal with) are required in these methods.

To overcome the above drawbacks, several efficient convolution methods are recently developed to design specific convolutional kernels for less parameters and calculations. In 2016, Szegedy et al. \cite{szegedy2016rethinking} proposed an asymmetrical convolution where a standard d$ \times $d convolution layer is spatially factorized as a sequence of two layers with d$ \times $1 and 1$ \times $d convolutions. Howard et al. \cite{howard2017mobilenets} proposed MobileNet v1 that replaces the standard convolution with the depth-wise separable convolution. The work by Zhang et al. \cite{zhang2018shufflenet} proposed ShuffleNet, applying group convolution and channel shuffle. Iandola et al. \cite{iandola2016squeezenet} proposed SqueezeNet in which 1$ \times $1 convolutions are utilized to reduce channel numbers and replace a part of 3$ \times $3 convolutions for less parameters. Although researches \cite{iandola2016squeezenet,zhang2018shufflenet,ma2018shufflenet} have investigated on reducing the channel number in the current layer to cut down the following convolutional operations, this problem is simply solved by appending 1$ \times $1 convolutional layer, which introduces extra parameters and considerable interchannel calculations.

In this paper, we found that feature redundancy exists among channels in CNN architecture, i.e., amounts of interchannel information is unimportant or even unnecessary in some cases. Instead of 1$ \times $1 convolutions, a novel convolutional construction named compact convolution is proposed to implicitly reduce feature redundancy in a non-learning approach. Specifically, the point-wise operation among channels (the point-wise interchannel operation) is implemented to squeeze the channel dimension of input feature maps. The reason for applying the point-wise operation is threefolds. First, the point-wise operation compresses the interchannel information without extra parameters, directly reducing the cost of computation. Second, the derivation of these point-wise operations can be taken easily, which contributes to the chain rule and training end-to-end networks from scratch. Third, the point-wise operation is well-suited for parallel computation on GPU or other advanced chips. Depth-wise separable convolution is further introduced to decouple spatial feature extraction from interchannel feature extraction. Like other research on efficient convolutional kernels\cite{howard2017mobilenets,sandler2018mobilenetv2,zhang2018shufflenet,iandola2016squeezenet}, useful features from feature maps can be extracted with fewer parameters and operations by simply replacing the standard convolution with our compact convolution. In addition, how different types of point-wise operations impact on interchannel feature compression is further investigated. While there is tremendous difference between sounds and images, our compact convolution yields desired performance in multiple tasks, such as acoustic scene classification, sound event detection and image classification. To the best of our knowledge, there is few work to verify the generalization of their models in across multiple medias.

Extensive experiments show that compared with general network constructions (such as VGG, Resnet and MobileNets), the network with compact convolutions (hereafter CompactNet) not only greatly reduces computation complexity, but also yields desirable performance. To further illustrate the difference between linear manner and non-linear one, three different point-wise operations were compared. Some guidelines are provide for investigating model compression and accerleration.

The contributions of this work are summarized as follows: 

1) A novel convolution named compact convolution is proposed to implicitly reduce feature redundancy in a non-learning approach. Different from the existing channel compression method which directly utilizes 1x1 convolution, the proposed compact convolution adopts the point-wise interchannel operation to squeeze the channel dimension of feature maps with no extra parameters. It turns out that compact convolutions not only cost 18 times less computation than standard convolutions in terms of 3$ \times $3 size, but also yield competitive performance.

2) Some guidelines on replacing learnable parameters and complex operations in convolutional layers are summarized. This facilitates further investigation on feature dimension reduction in CNNs.

3) The proposed convolution can be easily applied in general CNN architectures, by replacing the current convolutions with our compact convolutions. Moreover, the compact convolution can extract either audio or visual features to solve multimedia problems.

The reminder of the paper is organized as follows. Section II provides a brief survey of related work. Section III first presents the proposed compact convolution, and then applies it into several popular CNN architectures. In Section IV, extensive experiments are conducted to evaluate CompactNets. Finally, several conclusions and possible future works are given in Section V.

\section{Related work}

\subsection{1$ \times $1 convolution}
1$ \times $1 convolution was first proposed by Lin et al.\cite{lin2013network} as a universal function approximator for feature extraction on the local patches. They found that 1$ \times $1 convolution not only has great capability in modeling various distributions of latent concepts, but also facilitates the learnable interactions of cross-channel information. Sequent work in \cite{Szegedy_2015_CVPR,he2016deep} utilized 1$ \times $1 convolution for tuning the number of feature maps in CNN architecture. However, 1$ \times $1 convolution involves considerable parameters and operations. This work applied the point-wise interchannel operation to reduce the dimension of feature maps.

\subsection{Maxout function}
In 2013, Goodfellow et al.\cite{goodfellow2013maxout} proposed a maxout construction that performs a max pooling across multiple affine feature maps. It turns out that the maxout construction results in a piecewise linear function which is capable of modeling any convex function. Wu et al. \cite{wu2018light} proposed the Max-Feature-Map (MFM) layer as a variation of maxout activation to suppress low-activation neurons in each layer. Rather than a better function approximator, this paper focuses on the efficient approaches for reducing the interchannel redundancy, and compressing the dimension of feature maps in a larger range. Moreover, besides the max pooling, two more operations are investigated and further integrated into the proposed convolutional layer.

\subsection{Depth-wise separable convolution}
Howard et al.\cite{howard2017mobilenets} proposed MobileNets v1 which took the idea of the depth-wise separable convolution and achieved preferable results on small models. Depth-wise separable convolution consists of a depth-wise convolution for spatially filtering and a point-wise convolution (1$ \times $1 convolution) for exchanging information among channels. By replacing standard convolutions with depth-wise separable convolutions, the optimized network costs about 9 times less computation than the standard convolution at the cost of a small reduction in accuracy. Inspired by the depth-wise separable convolution, the compact convolution decouples spatial feature extraction from interchannel feature extraction. Moreover, the point-wise interchannel operation is introduced between the depth-wise convolution and 1$ \times $1 convolution. Thus, the efficiency of convolution is further improved.

\section{Proposed Method}

\subsection{The point-wise interchannel operation}
As shown in Fig. 1, the point-wise operation is implemented on the feature maps across channels. The input feature maps are firstly divided into groups. And a new feature map is extracted point by point over $ C $ feature maps in each group. Therefore, the parameter $ C $ can be deemed as a hyper-parameter for adjusting the ratio of channel compression. As $ C $ gets larger, the resulting construction becomes more compact.

\begin{figure}[!t]
	\centering
	\centerline{\includegraphics[width=8.5cm]{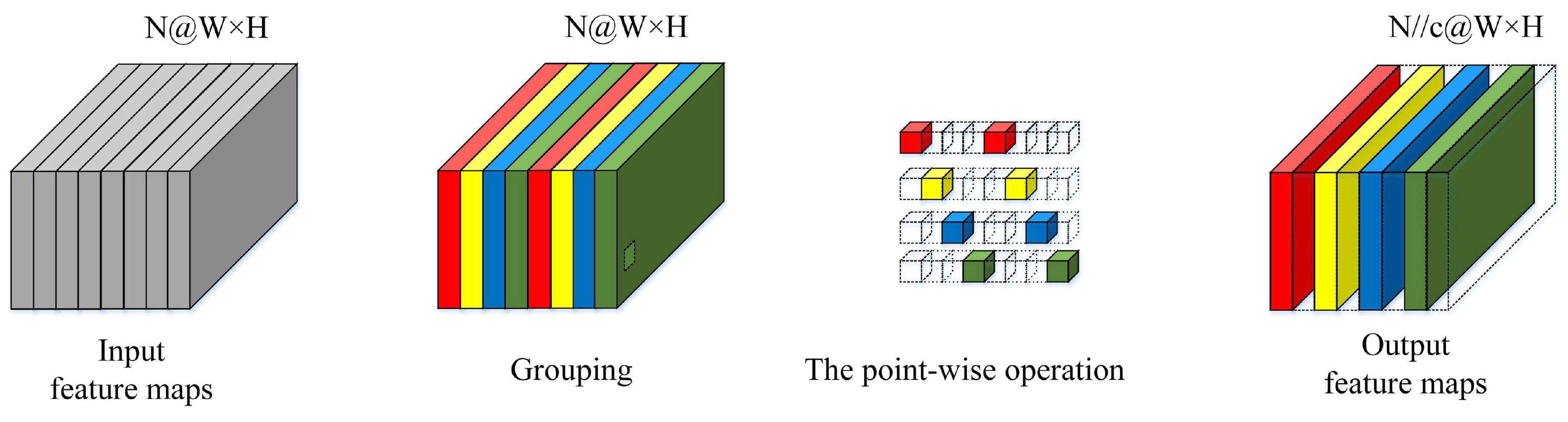}}
	\caption{Point-wise interchannel operation over feature maps. The compact factor $ C $ is set to 2 in this figure. Thus, the input feature maps are first groupedand then the point-wise operation are implemented over the 2 feature maps in each group.}
	\label{fig:1}
\end{figure}

The input feature maps and the output feature maps of the point-wise interchannel operation are denoted as $ I\in F^{N\times W\times H} $ and $ O\in F^{N’\times W\times H} $, where $ N $ and $ N’ $ are the channel numbers of input feature maps and output feature maps, $ W $ and $ H $ are the width and height of the feature maps respectively. Each pixel on the output feature maps are independently calculated with the values in the identical position across channels. Thus, the point-wise interchannel operation of the position $ \left( w, h \right)  $ $ \left( 0\leq w<W, 0\leq h<H\right)  $ is defined as

\begin{equation}
O_{w,h}(n)=T_{k=nC}^{(n+1)C-1}(I_{w,h}(k)),\quad n=0,1,\cdots,N'-1
\label{eqn:1}
\end{equation}

Here $ T(\ast) $ represents the point-wise operations across the channels ranged from $ nC $ to $ (n+1)C-1 $. The adopted point-wise operation can be divided into non-linear and linear manners. The non-linear manner which combines $ C $ feature maps and outputs element-wise maximum one is defined as:

\begin{equation}
O_{w,h}(n)=max_{k=nC}^{(n+1)C-1}(I_{w,h}(k)),\quad n=0,1,\cdots,N'-1
\label{eqn:2}
\end{equation}

The gradient of Eq. (2) takes the following form:

\begin{equation}
\frac{\partial O_{w,h}(n)}{\partial I_{w,h}(j)}= 
\begin{cases}
1,\quad \underset{nC\leq j \leq (n+1)C}{\arg\max\ }{I_{w, h}(j)}\\ 
0,\quad otherwise
\end{cases}
\label{eqn:3}
\end{equation}

Likewise, the linear manner is defined as:

\begin{equation}
\begin{split}
O_{w,h}(n)=&\frac{1}{m}[I_{w,h}(nC)+I_{w,h}(nC+1)+\dots\\
+&I_{w,h}((n+1)C-1)],\quad n=0,1,\cdots,N'-1
\end{split}
\label{eqn:4}
\end{equation}

Here $ m $ is set to 1 when the sum method is applied, otherwise set to $ C $.
The gradient of Eq. (4) can be written as follows:

\begin{equation}
\frac{\partial O_{w,h}(n)}{\partial I_{w,h}(j)}= 
\begin{cases}
1/m,\quad nC\leq j \leq (n+1)C\\ 
0,\quad otherwise
\end{cases}
\label{eqn:5}
\end{equation}

Because the point-wise operation can be simultaneously processed in different groups, it is well-suited for parallel computation on the modern processors. Compared with 1$ \times $1 convolution performing weighted linear recombination across all the input feature maps, each output feature map produced by the point-wise operation is calculated from the local information of the grouped input feature maps with no extra learnable weights. Thus, the point-wise interchannel operation is capable of reducing considerable parameters and computation resources.

\subsection{Compact convolution}

Taking advantages of the depth-wise separable convolution and the point-wise interchannel operation, a novel compact convolution layer is proposed for the efficient network. The proposed compact convolution is illustrated in Fig.2. Depth-wise convolution is operated over each input feature map to extract spatial features. The following point-wise interchannel operation squeezes the channel dimension of feature maps extracted by depth-wise convolutions, and maintains their major information. Finally, 1$ \times $1 convolution is applied for the exchange of information among channels. As one can see, there is a bottleneck construction inside the compact convolution. The bottleneck construction leaves the 1$ \times $1 layer with smaller input/output dimensions, which is benifit to less cost of computation. Compared with other bottleneck constructions\cite{he2016deep} designed with 1$ \times $1 convolution, the proposed compact convolution reduces the channel dimension with less calculation and no extra learnable weights.

\begin{figure}[htb]
	\centering
	\centerline{\includegraphics[width=8.5cm]{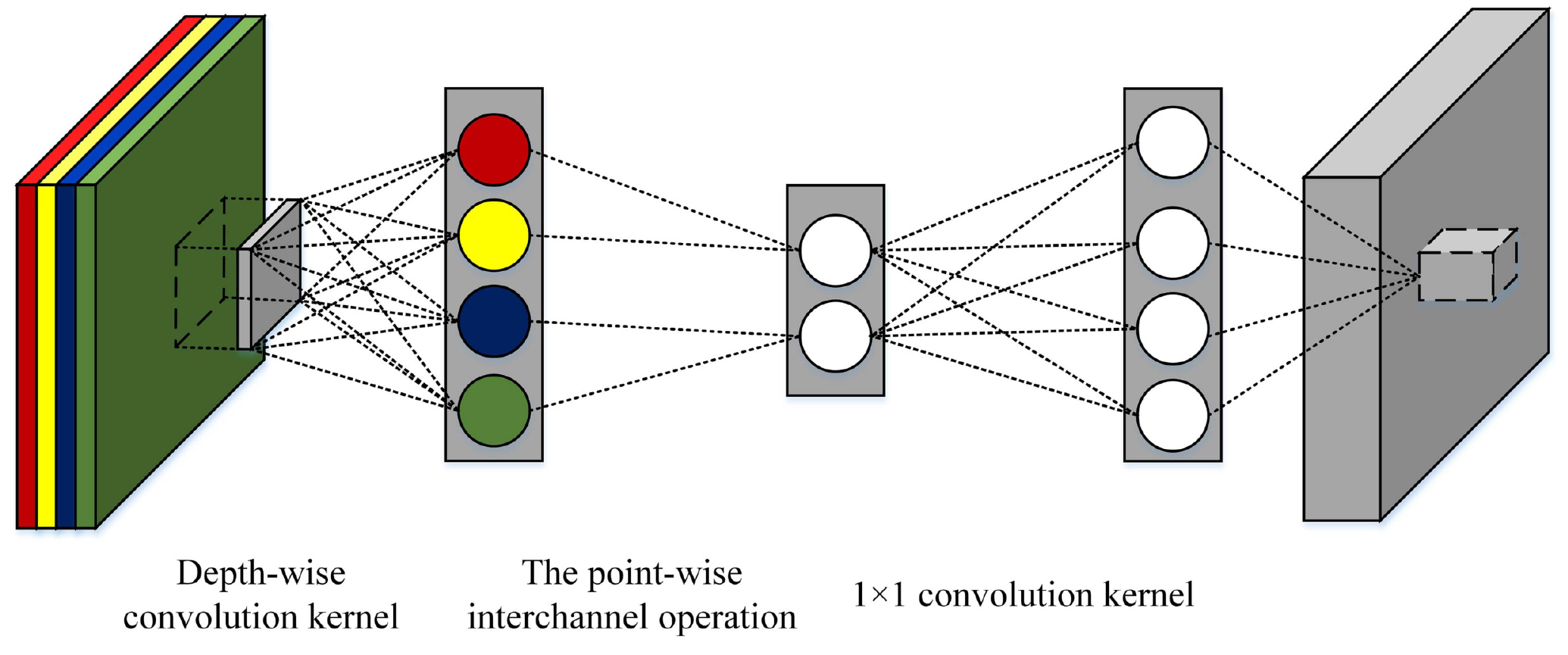}}
	\caption{Illustration of compact convolutions. Input feature maps go through depth-wise convolutions, point-wise interchannel operations and 1x1 convolutions. The compact factor \textit{C} is set to 2 in this figure. Different colors denote different channels of feature maps.}
	\label{fig:2}
\end{figure}

A standard convolution layer takes a $ W_{in}\times H_{in}\times C_{in} $ feature map $ F $ as input. Here $ W_{in} $ and $ H_{in} $ are the spatial width and height of the input feature map, $ C_{in} $ is the number of input channels. And a $ W_{out}\times H_{out} \times C_{out} $ feature map $ G $ is produced by a standard convolution, where $ W_{out} $ and $ H_{out} $ are the width and height of the output feature map and $ C_{out} $ is the number of output channel. The standard convolutional layer is parameterized by convolution kernel sized $ K\times K \times C_{in}\times C_{out} $ where $ K $ is the spatial dimension of the kernel assumed to be square, $ C_{in} $ and $ C_{out} $ are numbers of input and output channel as defined previously.

Based on\cite{molchanov2016pruning}, the complexities of networks are evaluated with FLOPs, i.e. the number of floating-point multiply-add operations. Assume that $ F $ denotes FLOPs of the standard convolution. It can be computed as:

\begin{equation}
F=2C_{in}K^2H_{out}W_{out}C_{out}
\label{eqn:6}
\end{equation}

Likewise, $ F’ $ represents FLOPs of the compact convolution. Through the depth-wise convolution, the point-wise interchannel operation and 1$ \times $1 convolution, $ F’ $ is calculated as:

\begin{equation}
F'=\left( 2K^2+C_{out}+m/2\right) C_{in}H_{out}W_{out}
\label{eqn:7}
\end{equation}
where $ m $ is set to 1 when the maximum and the sum methods are imposed, otherwise set to 2.
Then the compression rate $ \alpha $ of $ F’ $ over $ F $ is obtained as:

\begin{equation}
\alpha(F',F)=\frac{1}{C_{out}}+\frac{1}{2K^2}+\frac{m}{4K^2C_{out}}
\label{eqn:8}
\end{equation}

Since the standard convolution sized 3$ \times $3 is the most frequently-used construction in CNN architecture, the kernel size of the compact convolution is set to 3 in the experiments. It turns out that FLOPs of the compact convolution sized 3$ \times $3 are between 12 to 18 times less than FLOPs of the standard one at a slight decline in accuracy as further demonstrated in Sect. V.

\begin{figure*}[!t]
	\centering
	\subfloat[]{\includegraphics[width=3in]{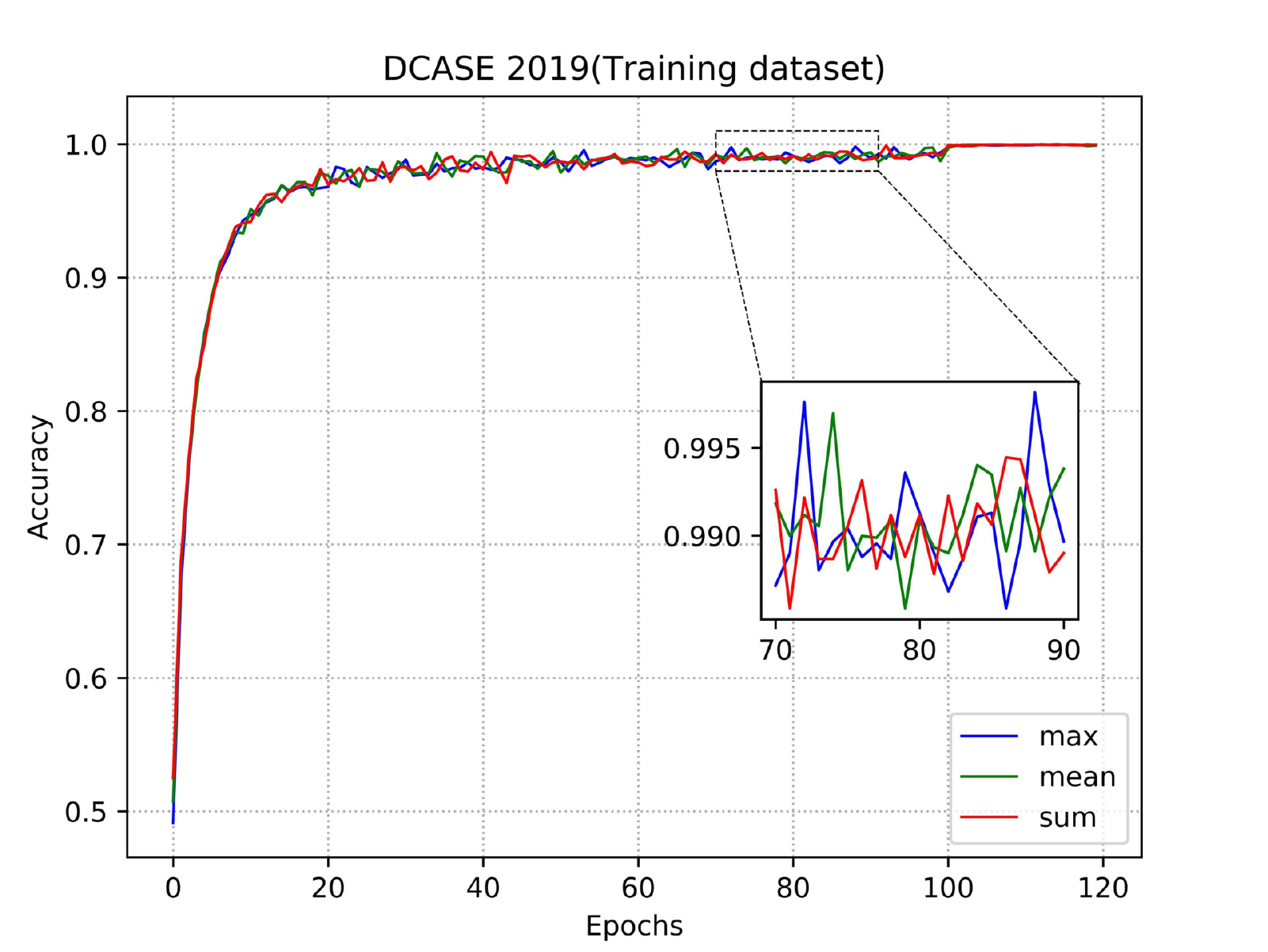}%
		\label{fig3a}}
	\hfil
	\subfloat[]{\includegraphics[width=3in]{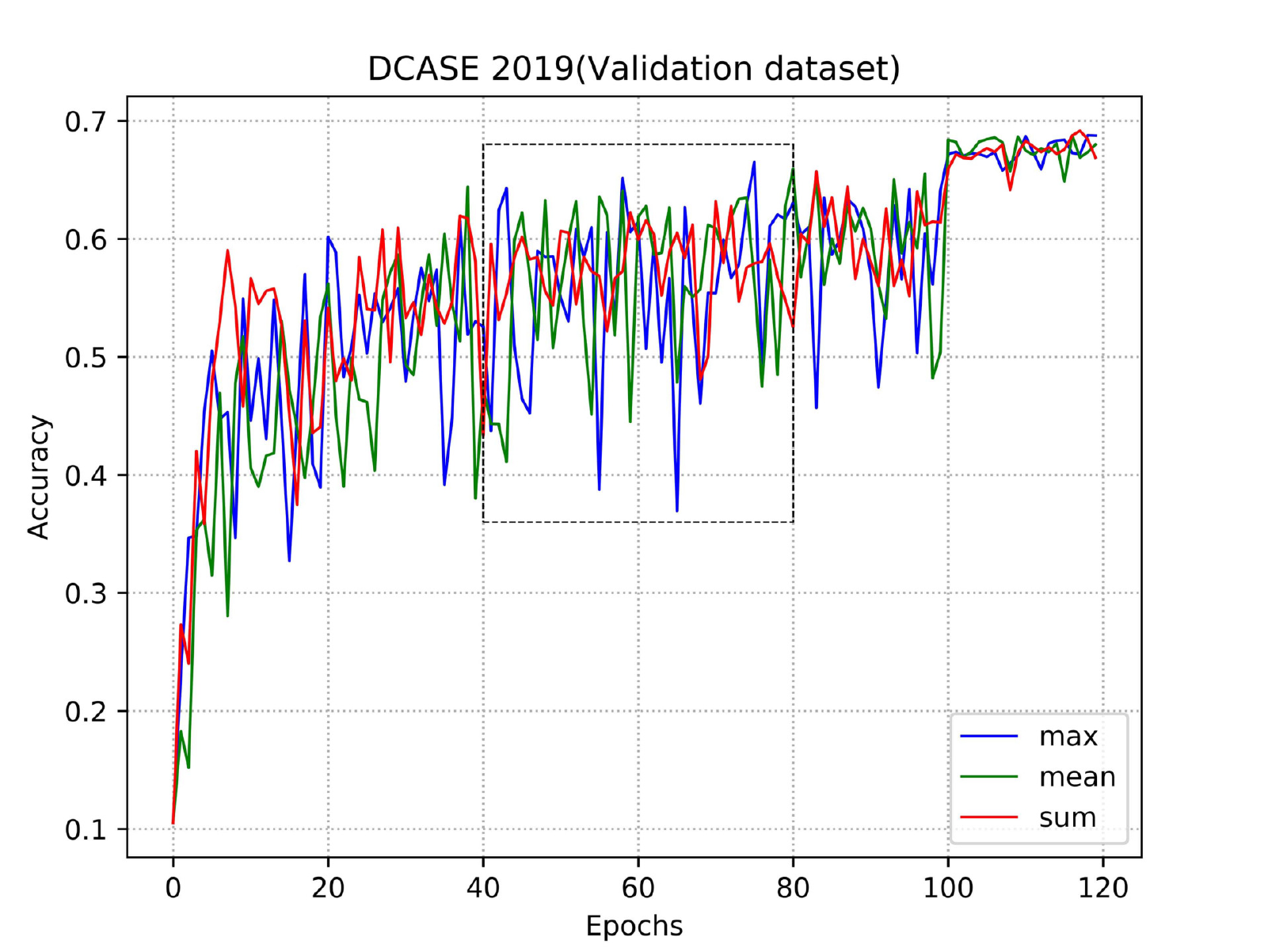}%
		\label{fig3b}}
	\hfil
	\subfloat[]{\includegraphics[width=3in]{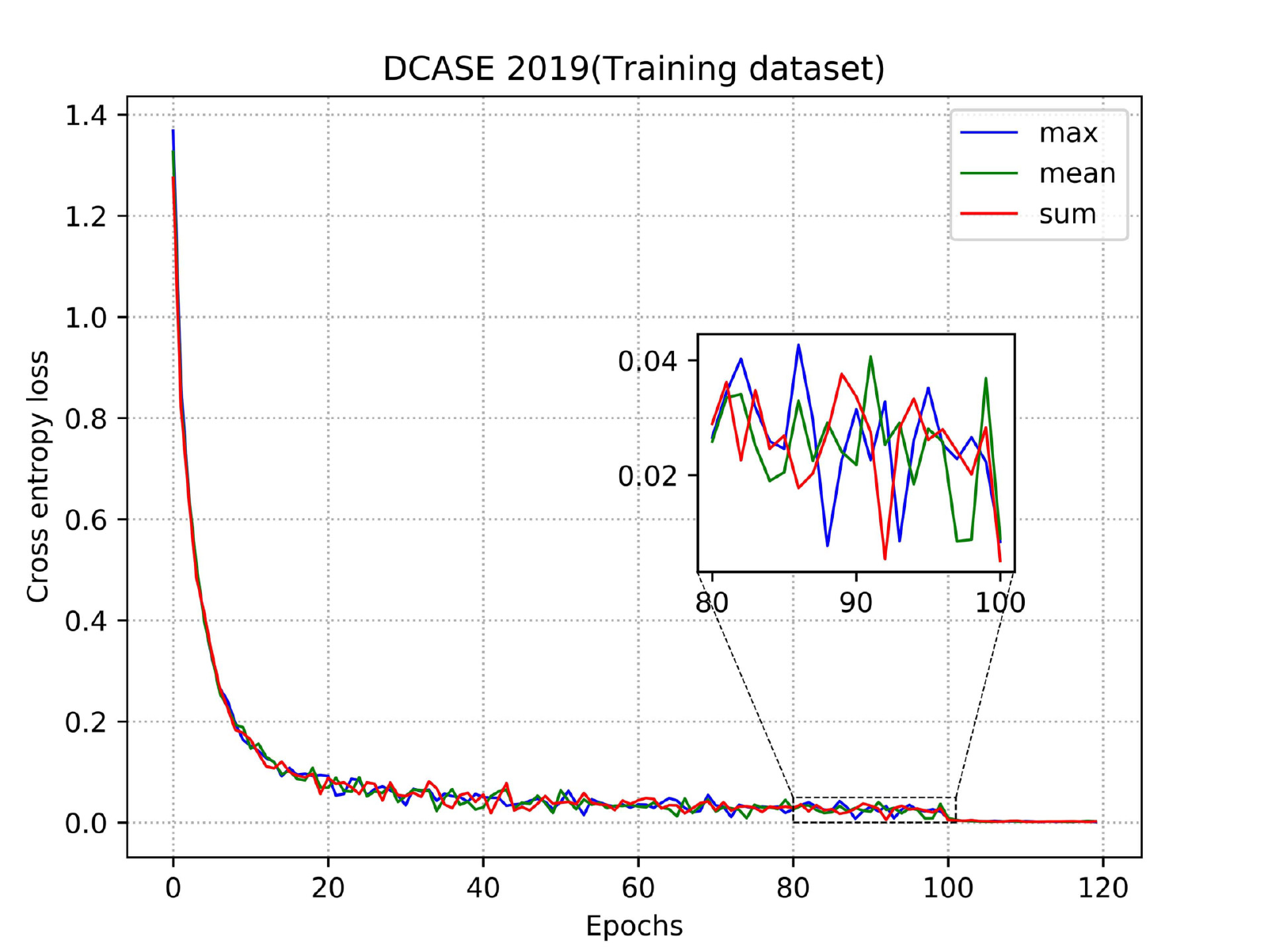}%
		\label{fig3c}}
	\hfil
	\subfloat[]{\includegraphics[width=3in]{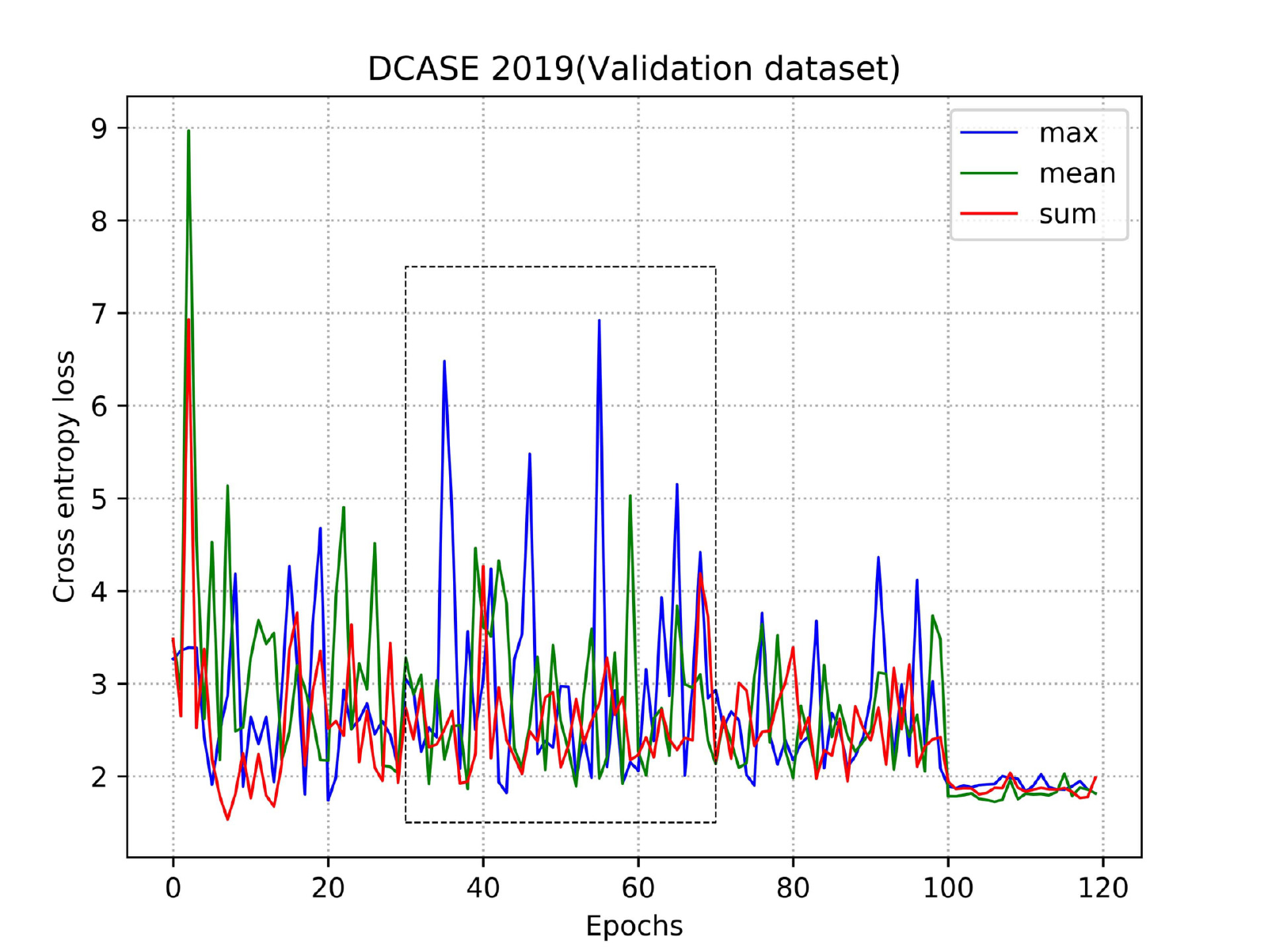}%
		\label{fig3d}}
	\hfil
	\caption{Accuracy and cross-entropy loss of three different point-wise interchannel operations on the DCASE 2019 dataset. (a) Training accuracy (b) Validation accuracy (c) Training loss (d) Validation loss}
	\label{fig3}
\end{figure*}

\subsection{Application in network architectures}
Since the compact convolution is a “sparse” version of the standard convolution, it can be embedded into general network architectures by simply replacing standard convolutions with compact convolutions. In this work, three different networks with compact convolutions are proposed as follows.

\textbf{VGG-like.} Following the design principle of VGG net\cite{simonyan2014very}, a block consisting of two-layer 3$ \times $3 convolutions is imposed as a basic building block. Considering the limitation of dataset size, an eight-layer stacked convolutional model is adopted in the proposed VGG-like networks. The standard convolution is utilized as the first two convolutional layers, and compact convolutions are imposed as the other six convolutional layers. All the convolutional layers are followed by batch normalization\cite{ioffe2015batch} and ReLU non-linear activation\cite{glorot2011deep}.

\textbf{ResNet-like.} The “bottleneck” design is adopted in our proposed networks, which has been demonstrated desired performance in \cite{he2016deep}. Different from \cite{he2016deep}, it is unnecessary to append 1$ \times $ 1 convolution following 3$ \times $3 convolution in the bottleneck block, because our compact convolution itself includes a 1$ \times $ 1 convolution.

\textbf{MobileNet-like.} To build MobileNet-like networks, depth-wise convolutions and point-wise convolutions are replaced with compact convolutions. In \cite{howard2017mobilenets}, the input channel number of a given depth-wise separable convolution with width multiplier $ \alpha $ is reduced from $ C_{in} $ to $ \alpha C_{in} $. Likewise, its output channel number is reduced to $ \alpha C_{out} $. Therefore, the model complexity with width multiplier $ \alpha $ decreases by roughly $ \alpha ^2 $. Since our compact convolution adjusts the number of channels through the point-wise interchannel operation, the width multiplier of the depth-wise convolution is fixed to 1 so as not to interfere with the experiments.

\subsection{Analysis in the training stage}

Different types of the point-wise operation make various impact among channels on both inference and backpropagation stage. In Eq. (4) and Eq. (5), except for weights, the sum and the average methods process the feature maps among channels in the same way. Therefore, the point-wise operation can be divided into linear and non-linear manner according to the interchannel processing. Empirically, the linear manner is prone to reserve the major information among local channels, while the non-linear one tends to extract prominent features among local channels. The accuracy and cross-entropy loss of three different point-wise interchannel operations on the DCASE 2019 dataset are shown in Fig. 3. The convergence of the max method is slower than the convergence of the other two methods on both the training dataset and  the validation dataset. In addition, it can be seen that the curves resulted by the sum and the average methods are similar, because both of them compress the information among channels in the linear manner.

\section{Generalization in multimedia}

To assess its capacity of generalization in cross media, the proposed networks are applied to tackle with three different tasks, including acoustic scene classification (ASC), sound event detection (SED) and image classification (IC). ASC and SED take 2-D time-frequency spectrograms as inputs to CNN classifier while IC directly utilizes images as inputs. “Acoustic scene” here is referred as a mixture of background noise and sound events associated with a specific audio scenario. So compared with SED, ASC tends to make the discrimination with more abstract and global features.

ASC aims at enabling devices to recognize the specific audio environment from a recording or an on-line stream. To solve this problem, the proposed networks are trained and evaluated on the development dataset of TAU Urban Acoustic Scenes 2019\cite{mesaros2018multi} in DCASE 2019 task 1. The dataset contains several acoustic scenes and various locations for each scene. The original recordings sampled with 44.1kHz are segmented into 10-second clips. The dataset consists of 10 scene classes, including airport, shopping mall, metro station, street pedestrian, public square, street traffic, tram, bus, metro and park.

To facilitate the proposed models training, the raw waves with binaural channels are firstly downmixed to mono. Then the log-scaled mel-spectrograms are extracted from each audio wave with hamming widow size of 1724 samples (corresponding to 0.04s), overlap of 50\%, and 128 mel bands. Therefore, a feature map with a size of 128$ \times $512 is generated for each audio waves. The features are finally normalized with z-scores, and fed into the proposed models.

SED aims to detect and classify events that occur in different environments. To solve this problem, the proposed networks are trained and evaluated on UrbanSound8K\cite{salamon2014dataset}. The dataset contains 8732 labeled sound clips of urban sounds from 10 classes, including air conditioner, car horn children playing, dog bark, drilling, engine idling, gun shot, jackhammer, siren and street music. Different from DCASE 2019, the length of clips is varying from 0s to 4s. The pre-processing on SED for training is similar with the one on ASC, except zero padding is adopted to unify the length of raw wave.

IC is a classical problem in computer vision. Aiming at evaluating the performance of our models on IC, CIFAR 10 is utilized for further experiments in Sect. V. CIFAR 10 contains 60000 32$ \times  $32 color images from 10 non-overlapping classes in the dataset, including airplane, automobile, bird, cat, deer, dog, frog, horse, ship and truck. Without much pre-processing, only normalization is applied for better convergence. The proposed networks are trained on 50000 samples, and validated on 10000 samples.

\begin{table*}[!t]
	\caption{Comparison of several models over parameters, complexity computations and speed on two platforms and three types of network architectures. The results of our CompactNets with the maximum, sum (left) and the average (right) methods are given independently. The speed on CPU and GPU are evaluated with single thread. The best results are highlighted in \textbf{bold}.}
	\label{tab:1}
	\centering
	\begin{tabular}{l|llll}
		\hline
		Model                           & Params     & Complexity (MFLOPs) & Speed on CPU (Samples/sec.) & Speed on GPU (Samples/sec.) \\ \hline
		VGG-8                           & 4,697,034  & 20233.6             & 1.70                        & 6.62                        \\
		XVGG-8                          & 580,362    & 6521.4              & 3.21                        & 7.41                        \\
		VGG-like CompactNet ($ C $=2)       & 329,034    & 5661.7/5663.9       & 3.04                        & 7.81                        \\
		VGG-like CompactNet ($ C $=4)       & 200,010    & 5231.8/5232.9       & 3.26                        & 7.94                        \\
		VGG-like CompactNet ($ C $=8)       & 135,498    & 5016.9/5017.4       & \textbf{3.32}               & \textbf{8.13}               \\ \hline
		ResNet                          & 23,601,930 & 9535.0              & 3.10                        & 5.59                        \\
		ResNet-like CompactNet ($ C $=2)    & 9,803,722  & 3860.2/3861.0       & 4.00                        & 6.21                        \\
		ResNet-like CompactNet ($ C $=4)    & 8,546,250  & 3341.9/3342.3       & 4.55                        & 6.33                        \\
		ResNet-like CompactNet ($ C $=8)    & 7,917,514  & 3082.7/3082.9       & \textbf{4.88}               & \textbf{6.37}               \\ \hline
		1.0 MobileNet v1                 & 3,238,538  & 1362.6              & 9.52                        & 62.50                       \\
		0.5 MobileNet v1                 & 834,378    & 354.1               & 11.36                       & 71.43                       \\
		0.25 MobileNet v1                & 220,970    & 95.3                & \textbf{20.83}              & \textbf{76.92}              \\
		MobileNet-like CompactNet ($ C $=2) & 1,668,746  & 709.5/710.8         & 10.31                       & 66.67                       \\
		MobileNet-like CompactNet ($ C $=4) & 883,850    & 382.9/383.6         & 12.35                       & 71.43                       \\
		MobileNet-like CompactNet ($ C $=8) & 491,402    & 219.7/220.0         & 14.08                       & \textbf{76.92}              \\ \hline  
	\end{tabular}
\end{table*}

\begin{table*}[!t]
	\caption{Comparison of several efficient convolutional neural networks over parameters, complexity computations and speed on two platforms. The best results are highlighted in \textbf{bold}.}
	\label{tab:2}
	\centering
	\begin{tabular}{l|llll}
		\hline
		Model                           & Params    & Complexity (MFLOPs) & Speed on CPU (Samples/sec.) & Speed on GPU (Samples/sec.) \\ \hline
		1.0 MobileNet v1                 & 3,238,538 & 1362.6              & 9.52                        & 62.50                       \\
		1.0 MobileNet v2                 & 2,288,458 & 700.4               & 6.99                        & 38.46                       \\
		ShuffileNet v1 2$\times$ ($ g $=3)          & 3,631,450 & 1221.2              & 6.71                        & 40.00                       \\
		ShuffileNet v2 2$\times$                & 5,407,316 & 1504.1              & 6.99                        & 58.82                       \\
		MobileNet-like CompactNet ($ C $=2) & 1,668,746 & 709.5/710.8         & \textbf{10.31  }                     & \textbf{66.67 }                      \\ \hline       
	\end{tabular}
\end{table*}

\section{Experimental Results}

\subsection{Experimental setup}
The proposed CompactNets with the sum, the max and the average methods are referred as CompactNet-S, CompactNet-M and CompactNet-A, respectively. Since compact convolution is applicable to most of the common network architectures, the proposed CompactNets are built in the same constructions as three different comparison networks, including VGG-8, ResNet and MobileNets. In addition, XVGG-8 is designed by replacing compact convolutions with depth-wise separable convolutions in order to evaluate the performance of the VGG-like CompactNet. To evaluate the efficiency of our CompactNets, some efficient convolutional neural networks (MobileNet v2, ShuffleNet v1 and Shufflenet v2) are built for comparison. Since nothing but convolutional layers changed in the following comparison experiments, only FLOPs of convolutions and our point-wise interchannel operations are taken into account. The above networks are trained by minimizing the cross-entropy loss with Adam optimizer. The learning rate, and batch size are set to 0.001 and 32 respectively.

All the experiments are implemented in python. Besides, experiments are conducted on the computer with Intel$ \textcircled R $Xeon(R) CPU E5-2650 v4 2.20 GHz and Nvidia RTX 2080Ti GPU. The proposed models are valued with Tensorflow.

\begin{table*}[!t]
	\caption{Accuracy of various models on DCASE 2019. The best results are highlighted in \textbf{bold}.}
	\label{tab:3}
	\centering
	\begin{tabular}{l|ll}
		\hline
		Model                             & Complexity (MFLOPs) & Accuracy (\%) \\ \hline
		VGG-8                             & 20233.6             & \textbf{69.96$ \pm  $0.31}   \\
		XVGG-8                            & 6521.4              & 68.56$ \pm  $1.06    \\
		VGG-like CompactNet-A ($ c $=2)       & 5663.9              & 68.30$ \pm  $0.75    \\
		VGG-like CompactNet-S ($ c $=2)       & 5661.7              & 68.66$ \pm  $1.44    \\
		VGG-like CompactNet-M ($ c $=2)       & 5661.7              & 68.88$ \pm  $0.92    \\ \hline
		ResNet                            & 9535.0              & 66.44$ \pm  $1.12    \\
		ResNet-like CompactNet-A ($ c $=2)    & 3861.0              & 68.78$ \pm  $0.40    \\
		ResNet-like CompactNet-S ($ c $=2)    & 3860.2              & \textbf{69.00$ \pm  $0.79}    \\
		ResNet-like CompactNet-M ($ c $=2)    & 3860.2              & 68.04$ \pm  $0.67    \\ \hline
		1.0 MobileNet v1                   & 1362.6              & 67.86$ \pm  $0.63    \\
		MobileNet-like CompactNet-A ($ c $=2) & 710.8               & 67.92$ \pm  $0.31    \\
		MobileNet-like CompactNet-S ($ c $=2) & 709.5               & 67.92$ \pm  $1.23    \\
		MobileNet-like CompactNet-M ($ c $=2) & 709.5               & \textbf{68.08$ \pm  $1.03}    \\ \hline
	\end{tabular}
\end{table*}

\begin{figure*}[!t]
	\centering
	\subfloat[]{\includegraphics[width=2.5in]{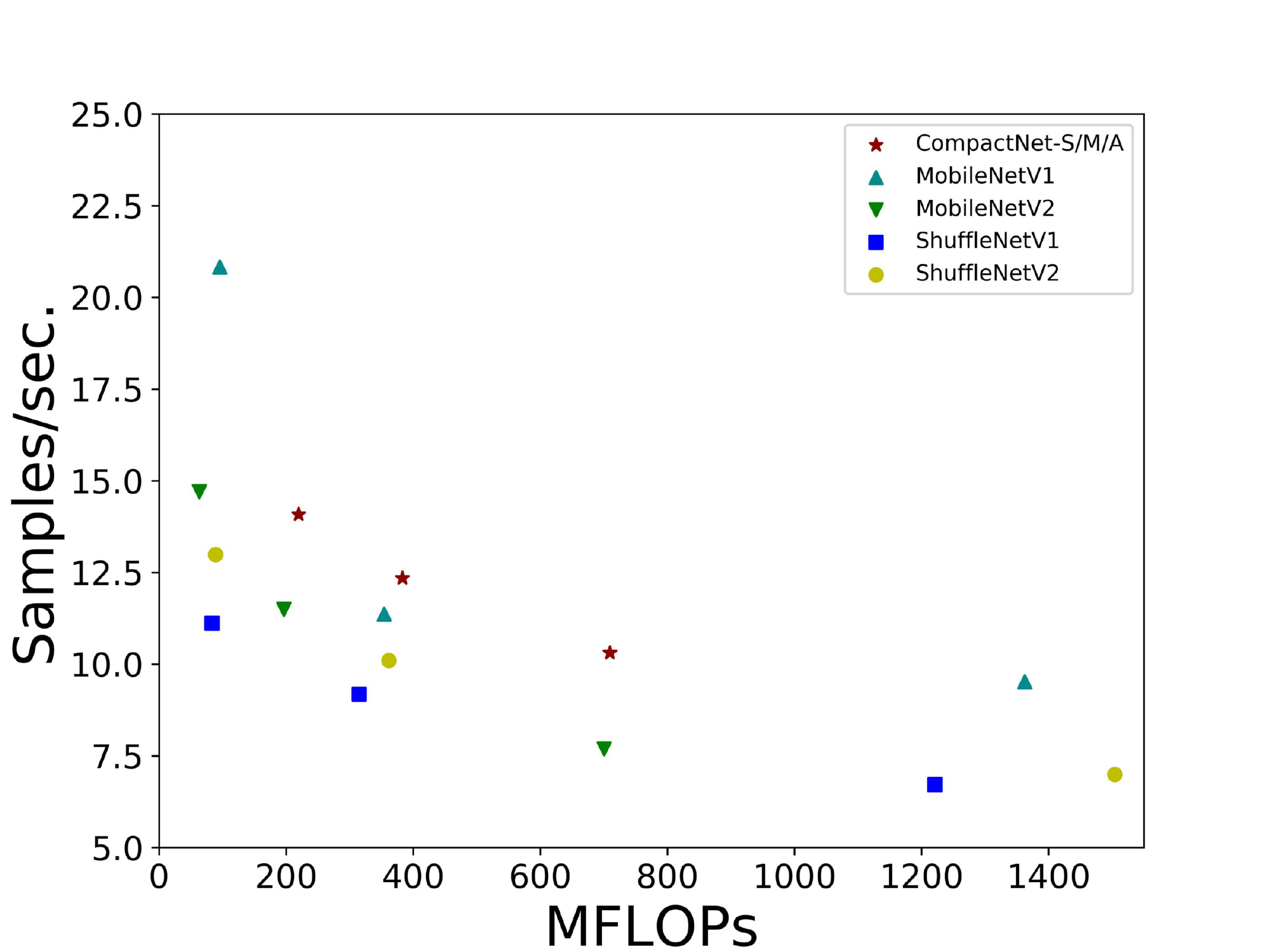}%
		\label{fig:4a}}
	\hfil
	\subfloat[]{\includegraphics[width=2.5in]{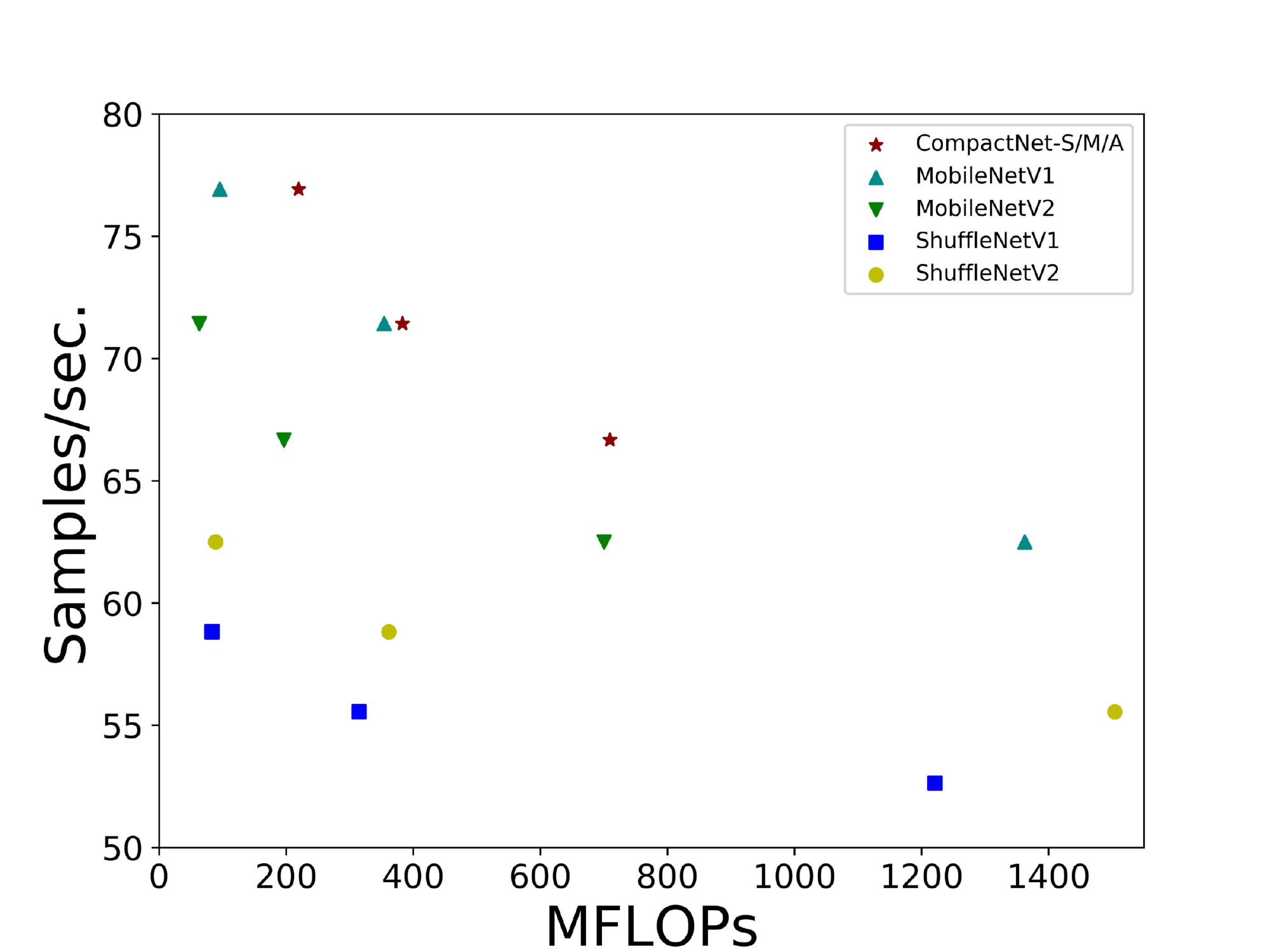}%
		\label{fig:4b}}
	\hfil
	\caption{Calculation complexity vs. speeds on two different platforms. (a) On CPU (b) On GPU.}
	\label{fig:4}
\end{figure*}

\begin{figure*}[!t]
	\centering
	\subfloat[]{\includegraphics[width=2.5in]{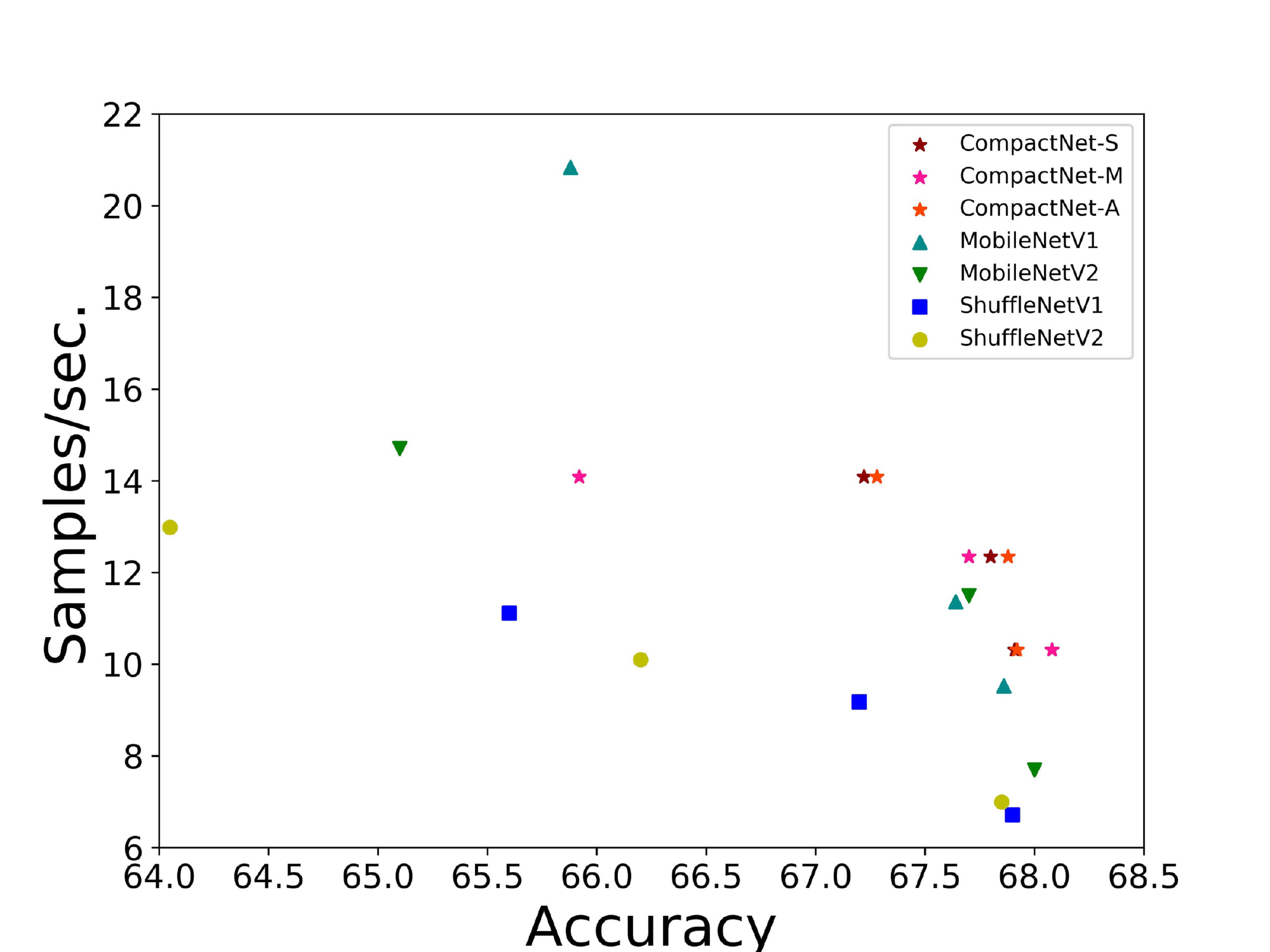}%
		\label{fig:5a}}
	\hfil
	\subfloat[]{\includegraphics[width=2.5in]{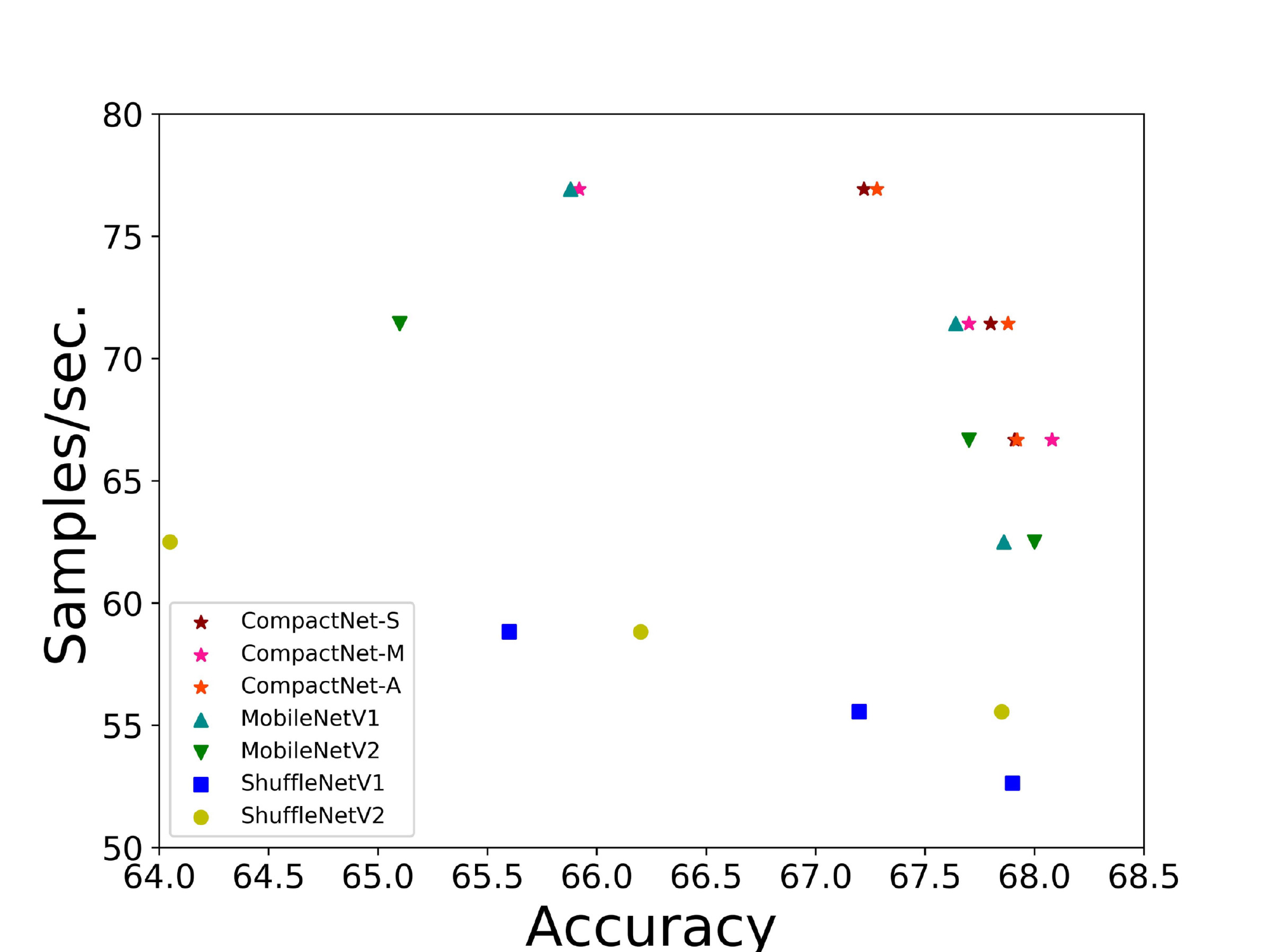}%
		\label{fig:5b}}
	\hfil
	\caption{Accuracy on DCASE 2019 vs. speeds on two different platforms. (a) On CPU (b) On GPU.}
	\label{fig:5}
\end{figure*}

\subsection{Algorithmic complexity}

The parameters, complexities and speeds of different models are listed in Table 1. The FLOPs of compact convolutions with the max, the sum and the average methods are given independently. For better observation, the results are grouped by different network architectures. Except the MobileNet-like networks on CPU, CompactNets ($ C $=8) are fastest on both CPU and GPU among the networks in the identical structures. Specifically, the VGG-like CompactNet ($ C $=8) speeds are 1.95$ \times $ and 1.23$ \times $ more than those of VGG-8 on CPU and GPU respectively. The ResNet-like CompactNet ($ C $=8) speeds are 1.57$ \times $ and 1.14$ \times $ more than those of ResNet on CPU and GPU respectively. It turns that 0.25 MobileNet v1 is faster than Mosbile-like CompactNets. This is because the complexity of 0.25 MobileNet v1 is merely a half of MobileNet-like CompactNet ($ C $=8) complexity. The non-linearity reduction of parameters and FLOPs are caused by the other unchanged convolutions in the networks, such as the first two standard convolutions in the VGG-like networks. Similarly, there are merely a few significant changes in complexity among the three proposed ResNet-like CompactNets, because only one 1$ \times $1 convolution at the end gets compacted while the other 1$ \times $1 convolutions have no change. 

Table 2 lists parameters, computation complexity and speeds of several efficient convolutional neural network. The speeds of mobileNet-like CompactNet ($ C $=2) are the fastest on both CPU and GPU.
Calculation complexity vs. speeds on two different platforms are shown in Fig. 4. Our proposed
CompactNets are on the top right region under both cases. It is worthy to note that the indirect metric (complexity) is inconsistent with the direct one (speed), e.g. the difference between CompactNet ($ C $=2) and 1.0 MobileNet v2. This result conforms to the finding in \cite{ma2018shufflenet}: Besides FLOPs, Memory access cost (MAC) and optimized operation on specific platforms should be also taken into consideration.

\begin{figure*}[!t]
	\subfloat[]{\includegraphics[width=0.5\linewidth]{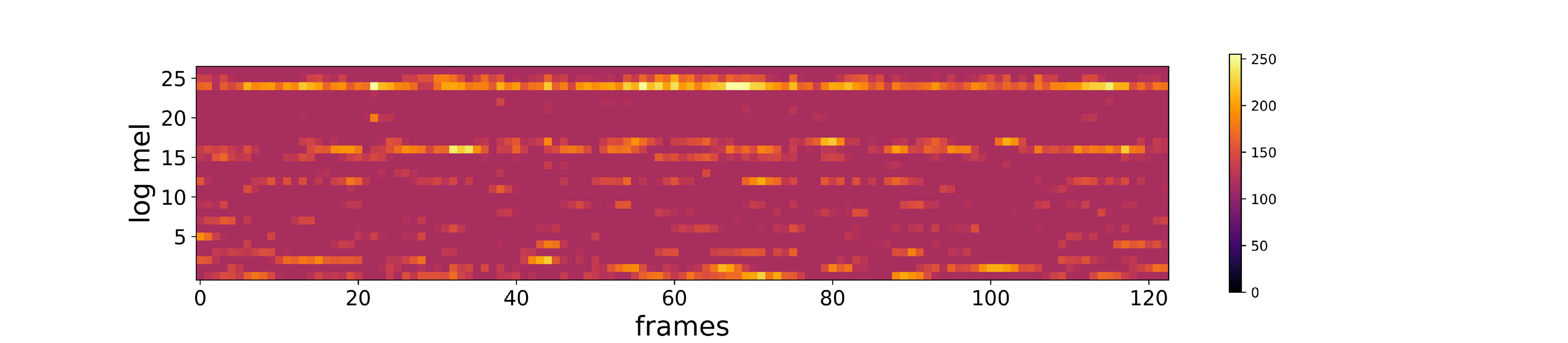}%
		\label{fig:6a}}
	\hfil
	\subfloat[]{\includegraphics[width=0.5\linewidth]{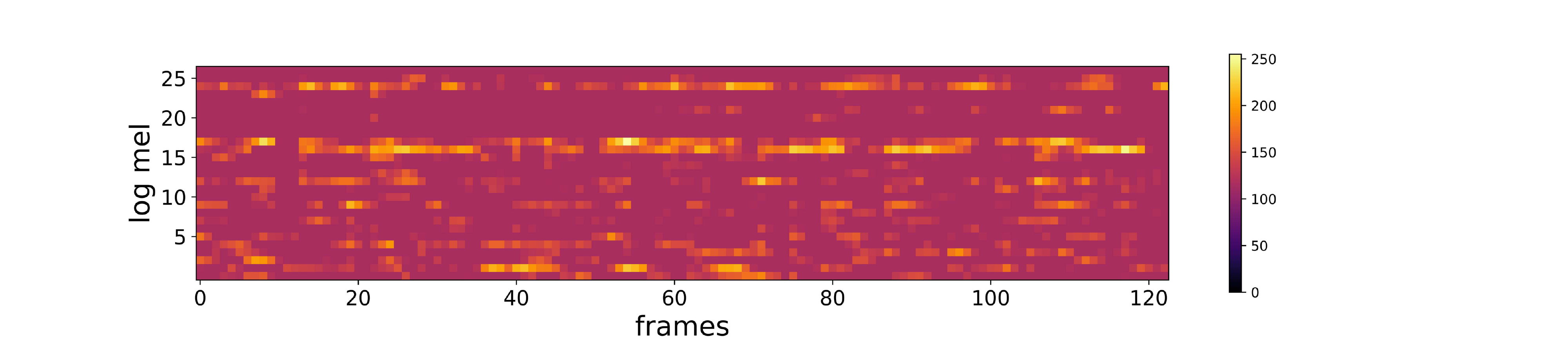}%
		\label{fig:6b}}
	\hfil
	\subfloat[]{\includegraphics[width=0.5\linewidth]{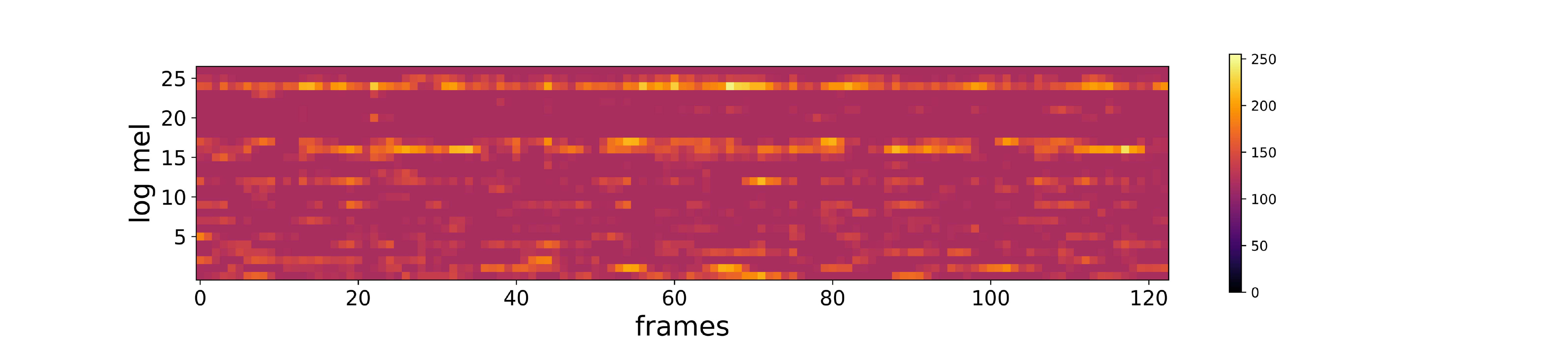}%
		\label{fig:6c}}
	\hfil
	\subfloat[]{\includegraphics[width=0.5\linewidth]{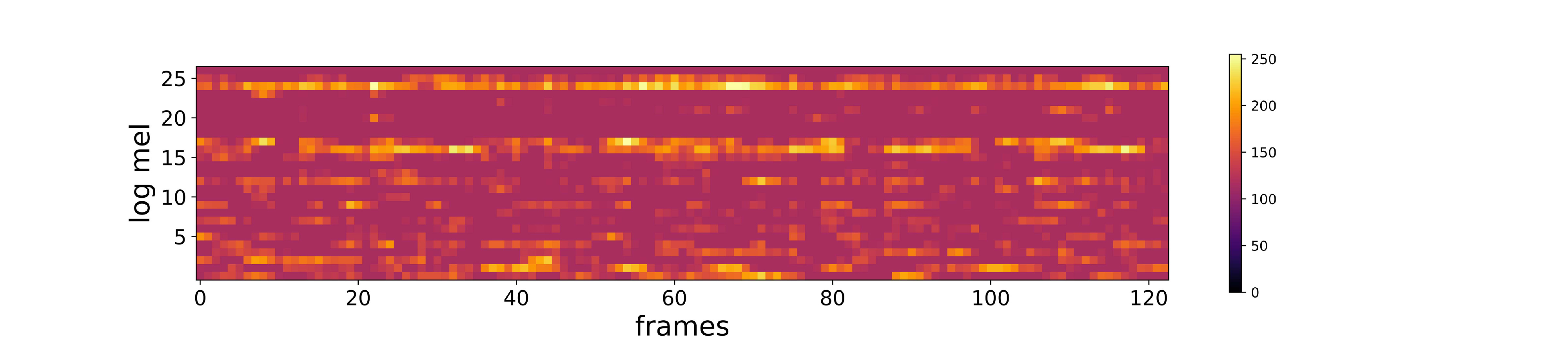}%
		\label{fig:6d}}
	\hfil
	\subfloat[]{\includegraphics[width=0.5\linewidth]{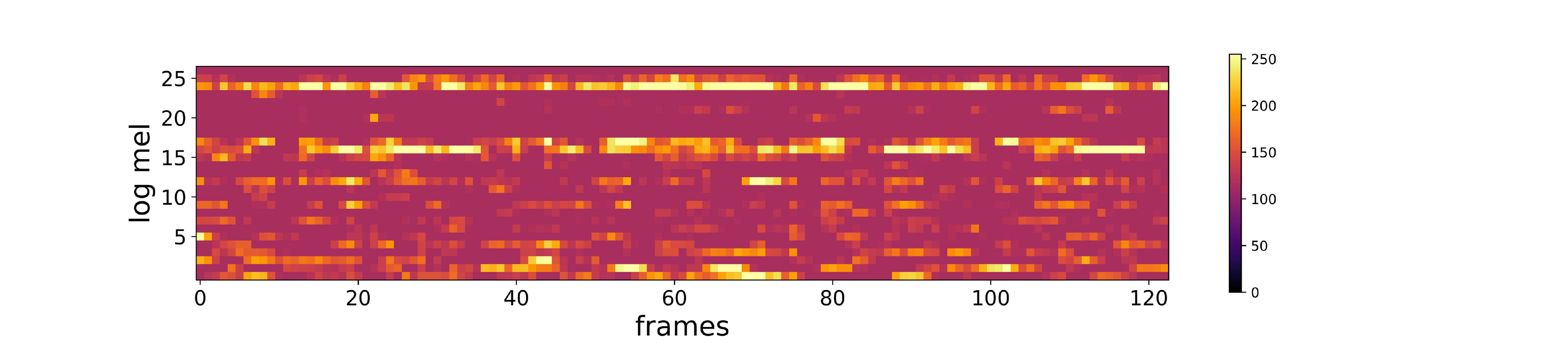}%
		\label{fig:6e}}
	\caption{Internal feature maps in CNN architecture. (a), (b) are the two input feature maps in the same group. (c), (d), (e) are the compact results with the average, the max and the sum methods respectively. The horizontal axis corresponds to the temporal frames limited to 10s, and the vertical axis corresponds to logarithmic mel-frequency bands. The color in the spectrograms reflects the energy intensity.}
	\label{fig:6}
\end{figure*}

\subsection{Evaluation on ASC}

Table 3 shows the accuracy of different models to handle ASC task on DCASE 2019. The proposed CompactNet-S and CompactNet-M yield the best results among ResNet-like models and MobileNet-like models respectively. It can be seen that VGG-8 outperforms the CompactNets by 1.3\%. However, taking the model complexity into consideration, the proposed models are still competitive. Compared with XVGG-8 and MobileNet v1 consisting of separable convolution, CompactNet-M still outperforms them by 0.32\% and 0.22\% respectively. This indicates that the point-wise interchannel operation can not only squeeze the channel dimension of input feature maps but also filter the useful information which helps further feature extraction. Note that the CompactNets surpass the ResNet by a large margin, because the ResNet is overfitting due to the limitation of dataset. Thus, the proposed compact convolution is capable of avoiding overfitting by reducing the number of learnable weights.

The accuracy on DCASE 2019 vs. the speeds on two different platforms are demonstrated on Fig. 5. Our proposed CompactNets are on the top right region under both cases. It turns out that the performance of comparison networks, such as ShuffleNet v1 and ShuffleNet v2, deteriorates rapidly along with the decrease of the scale factor. In contrast, when the compact factor $ C $ increases, the variation of our CompactNet accuracy is small. This indicates that the point-wise interchannel operation can squeeze the channel dimension of feature maps while retain the useful information in features.

\begin{figure}[!t]
	\centering
	\includegraphics[width=2.5in]{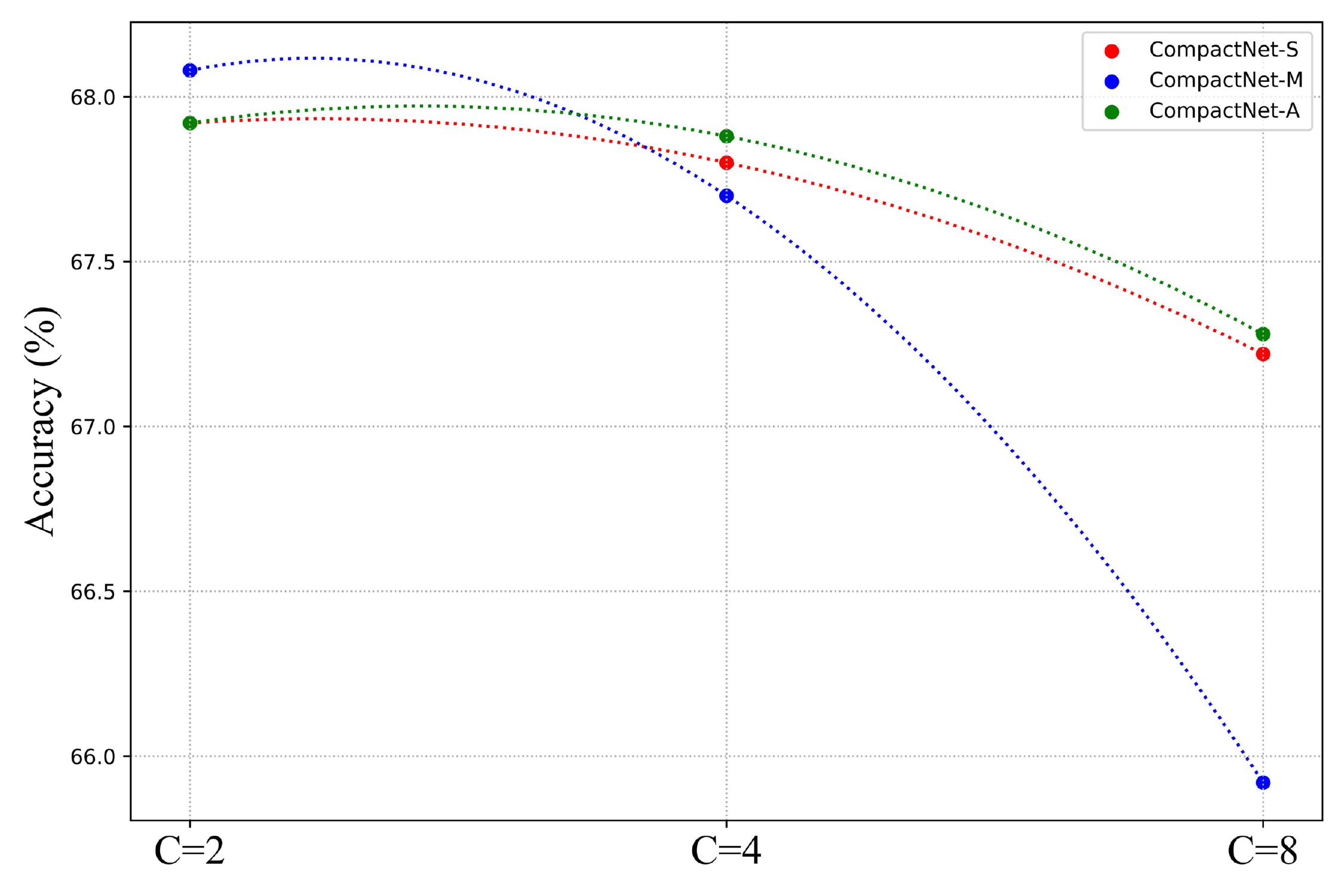}
	\caption{The accuracy variations of CompactNets with three different point-wise interchannel operations.}
	\label{fig:7}
\end{figure}

\begin{table*}[!t]
	\caption{Comparison of several models over complexity computations and accuracy in two different tasks.}
	\label{tab:4}
	\centering
	\begin{tabular}{l|ll|ll}
		\hline
		\multirow{2}{*}{Model}            & \multicolumn{2}{l|}{Sound Event Detection (SED)} & \multicolumn{2}{l}{Image Classification (IC)} \\ \cline{2-5}
		& Complexity (MFLOPs)         & acc. (\%)         & Complexity (MFLOPs)        & acc. (\%)        \\ \hline
		VGG-8                             & 7943.3                      & 84.34$ \pm  $1.41        & 15549.2                    & 87.58$ \pm  $0.16       \\
		XVGG-8                            & 2560.2                      & 81.42$ \pm  $1.90        & 5050.8                     & 85.18$ \pm  $0.23       \\
		VGG-like CompactNet-S ($ C $=2)       & 2222.7                      & 82.74$ \pm  $1.03        & 4392.5                     & 84.12$ \pm  $0.13       \\
		VGG-like CompactNet-M ($ C $=2)       & 2222.7                      & 83.34$ \pm  $1.43        & 4392.5                     & 83.20$ \pm  $0.30       \\
		VGG-like CompactNet-S ($ C $=4)       & 2053.9                      & 82.20$ \pm  $1.94        & 4063.4                     & 82.76$ \pm  $0.27       \\
		VGG-like CompactNet-M ($ C $=4)       & 2053.9                      & 82.06$ \pm  $1.22        & 4063,4                     & 80.80$ \pm  $0.28       \\
		VGG-like CompactNet-S ($ C $=8)       & 1969.5                      & 80.94$ \pm  $1.35        & 3898,8                     & 80.64$ \pm  $0.13       \\
		VGG-like CompactNet-M ($ C $=8)       & 1969.5                      & 80.06$ \pm  $1.39        & 3898,8                     & 77.10$ \pm  $0.39       \\ \hline
		ResNet                            & 3743.2                      & 77.82$ \pm  $0.88        & 7614.9                     & 72.72$ \pm  $5.03       \\
		ResNet-like CompactNet-S ($ C $=2)    & 1515.4                      & 79.40$ \pm  $1.04        & 3270.2                     & 74.20$ \pm  $1.41       \\
		ResNet-like CompactNet-M ($ C $=2)    & 1515.4                      & 78.08$ \pm  $0.38        & 3270.2                     & 73.70$ \pm  $0.83       \\
		ResNet-like CompactNet-S ($ C $=4)    & 1311.9                      & 79.18$ \pm  $1.06        & 2873,3                     & 73.90$ \pm  $1.26       \\
		ResNet-like CompactNet-M ($ C $=4)    & 1311.9                      & 78.02$ \pm  $0.94        & 2873.3                     & 72.68$ \pm  $0.67       \\
		ResNet-like CompactNet-S ($ C $=8)    & 1210.2                      & 79.02$ \pm  $1.40        & 2674.9                     & 73.72$ \pm  $0.46       \\
		ResNet-like CompactNet-M ($ C $=8)    & 1210.2                      & 77.84$ \pm  $1.67        & 2674.9                     & 71.34$ \pm  $1.15       \\ \hline
		1.0 MobileNet v1                   & 534.9                       & 77.50$ \pm  $1.06        & 1050.5                     & 78.00$ \pm  $0.37       \\
		MobileNet-like CompactNet-S ($ C $=2) & 278.5                       & 78.42$ \pm  $1.09        & 550.4                      & 76.80$ \pm  $0.34       \\
		MobileNet-like CompactNet-M ($ C $=2) & 278.5                       & 78.46$ \pm  $0.68        & 550.4                      & 75.00$ \pm  $0.32       \\
		MobileNet-like CompactNet-S ($ C $=4) & 150.3                       & 77.88$ \pm  $1.25        & 300.4                      & 74.22$ \pm  $0.54       \\
		MobileNet-like CompactNet-M ($ C $=4) & 150.3                       & 77.30$ \pm  $0.67        & 300.4                      & 71.82$ \pm  $0.80       \\
		0.5 MobileNet v1                   & 139.0                       & 75.58$ \pm  $1.28        & 274.8                      & 73.36$ \pm  $0.74       \\
		MobileNet-like CompactNet-S ($ C $=8) & 86.3                        & 77.58$ \pm  $2.46        & 175.4                      & 70.14$ \pm  $1.15       \\
		MobileNet-like CompactNet-M ($ C $=8) & 86.3                        & 75.38$ \pm  $2.06        & 175.4                      & 68.12$ \pm  $0.54       \\
		0.25 MobileNet v1                  & 37.4                        & 73.30$ \pm  $1.95        & 74.8                       & 66.92$ \pm  $0.91       \\ \hline
	\end{tabular}
\end{table*}

\subsection{Comparison between linear and non-linear manners}

Fig. 6 illustrates the internal feature maps resulting in the three different point-wise interchannel operations. The max method is clearer than the average method in the detailed information. This indicates that the max method can extract the iconic features from inputs while the average method tends to restore the major information in feature maps. In addition, the distribution of feature with the average method is identical to the one with the sum method. This phenomenon accords to the analysis in Sect. III \textit{A} and \textit{B}.

In Fig. 7, the accuracy variations of CompactNets with three different point-wise interchannel operations are illustrated. With the increase of compact factor $ C $, the performance of CompactNet-S always consistents with the performance of CompactNet-A. Combining the analysis in Sect. III \textit{A} and \textit{D}, we can summarize several guidelines: 

\textbf{G1) the average method and the sum method among channels work in the same way.} By taking FLOPs of these two methods into consideration, the average operation can be replaced with the sum operation to squeeze the channel dimension of input feature maps. 

\textbf{G2) The non-linear operation is relatively hard to convergence, and it tends to yield desirable performance with small compact factors.} The maximum method extracts the maximum value within a group and discards the remaining ones. As the compact factor $ C $ gets large, this nonlinear mapping loses a large amount of characteristic information, which leads to a rapid deterioration in performance.

\textbf{G3) The linear operation is relatively easy to convergence, and it tends to outperform other methods in the case of large compact factors.} In contrast to maximum method, average and sum methods preserve most of the information by arithmetic averaging. This facilitates model compression with a large compact factor.

These three guidelines can not only help researchers utilize CompactNets, but also expose the role of different operations in CNNs.

\subsection{Extend to other tasks}

Based on \textbf{G1}, only the sum method and the max method, corresponding to linear manner and non-linear one respectively, are discussed in this subsession. 

Table 4 lists the computation complexity and the accuracy in two different tasks. It turns out that our proposed CompactNets produce satisfying results in SED and IC. In SED, our CompactNet-S (($ C $=2) and CompactNet-M (($ C $=2) surpass the competing models among ResNet-like models and MobileNet-like models by 1.58\% and 0.96\% respectively. Compared with XVGG-8 that consists of separable convolutions, CompactNet-M (($ C $=2) still conducts higher accuracy by 1.92\%. In IC, CompactNet-S ($ C $=2) outperforms ResNet by about 1.48\%. It is worth to note that XVGG-8 and 1.0 MobileNet v1 yield better results than CompactNets by 1.06\% and 1.2\%. This is because the number of samples in CIFAR 10 is large, and each sample sized 32$\times$32 is easy to learn. Therefore, the input feature maps have less leeway to be squeezed.

\section{Conclusion}
In this paper, a novel convolutional construction was proposed for implicitly reducing feature redundancy, where the point-wise interchannel operation was adopted to squeeze the number of channel of feature maps. The depth-wise separable convolution and the point-wise interchannel operation were integrated to speed up calculations and retain a satisfying performance. Unlike traditional methods for dimensional reduction in CNN which introduce considerable learnable weights, our compact convolution has the capacity to squeeze the channel dimension of feature maps with no extra parameters. Moreover, we showed the capacity of generalization to handle three different tasks, including acoustic scene classification, sound event detection and image classification. Extensive experimental results demonstrated that the proposed method can not only cut down the run time on CPU and GPU but also produce promising performance.

In future, we will investigate proper alternatives to the current convolutional construction with less complexity, and applications to other general multimedia tasks.


%

%



\ifCLASSOPTIONcaptionsoff
  \newpage
\fi



%
%

\bibliographystyle{IEEEtran}
\bibliography{main}

\begin{thebibliography}{10}
\providecommand{\url}[1]{#1}
\csname url@samestyle\endcsname
\providecommand{\newblock}{\relax}
\providecommand{\bibinfo}[2]{#2}
\providecommand{\BIBentrySTDinterwordspacing}{\spaceskip=0pt\relax}
\providecommand{\BIBentryALTinterwordstretchfactor}{4}
\providecommand{\BIBentryALTinterwordspacing}{\spaceskip=\fontdimen2\font plus
\BIBentryALTinterwordstretchfactor\fontdimen3\font minus
  \fontdimen4\font\relax}
\providecommand{\BIBforeignlanguage}[2]{{%
\expandafter\ifx\csname l@#1\endcsname\relax
\typeout{** WARNING: IEEEtran.bst: No hyphenation pattern has been}%
\typeout{** loaded for the language `#1'. Using the pattern for}%
\typeout{** the default language instead.}%
\else
\language=\csname l@#1\endcsname
\fi
#2}}
\providecommand{\BIBdecl}{\relax}
\BIBdecl

\bibitem{8371638}
S.~{Xie} and H.~{Hu}, ``Facial expression recognition using hierarchical
  features with deep comprehensive multipatches aggregation convolutional
  neural networks,'' \emph{IEEE Transactions on Multimedia}, vol.~21, no.~1,
  pp. 211--220, Jan 2019.

\bibitem{8669681}
H.~{Cao}, H.~{Liu}, E.~{Song}, G.~{Ma}, X.~{Xu}, R.~{Jin}, T.~{Liu}, and
  C.~{Hung}, ``Multi-branch ensemble learning architecture based on 3d cnn for
  false positive reduction in lung nodule detection,'' \emph{IEEE Access},
  vol.~7, pp. 67\,380--67\,391, 2019.

\bibitem{8735820}
C.~{Li} and B.~{Yang}, ``Adaptive weighted cnn features integration for
  correlation filter tracking,'' \emph{IEEE Access}, vol.~7, pp.
  76\,416--76\,427, 2019.

\bibitem{8922774}
J.~{Xie}, K.~{Hu}, M.~{Zhu}, J.~{Yu}, and Q.~{Zhu}, ``Investigation of
  different cnn-based models for improved bird sound classification,''
  \emph{IEEE Access}, vol.~7, pp. 175\,353--175\,361, 2019.

\bibitem{ZHANG2020113067}
\BIBentryALTinterwordspacing
T.~Zhang, J.~Liang, and B.~Ding, ``Acoustic scene classification using deep cnn
  with fine-resolution feature,'' \emph{Expert Systems with Applications}, vol.
  143, p. 113067, 2020. [Online]. Available:
  \url{http://www.sciencedirect.com/science/article/pii/S0957417419307845}
\BIBentrySTDinterwordspacing

\bibitem{7952132}
S.~{Hershey}, S.~{Chaudhuri}, D.~P.~W. {Ellis}, J.~F. {Gemmeke}, A.~{Jansen},
  R.~C. {Moore}, M.~{Plakal}, D.~{Platt}, R.~A. {Saurous}, B.~{Seybold},
  M.~{Slaney}, R.~J. {Weiss}, and K.~{Wilson}, ``Cnn architectures for
  large-scale audio classification,'' in \emph{2017 IEEE International
  Conference on Acoustics, Speech and Signal Processing (ICASSP)}, March 2017,
  pp. 131--135.

\bibitem{8734068}
Y.~{Jin}, C.~{Luo}, W.~{Guo}, J.~{Xie}, D.~{Wu}, and R.~{Wang}, ``Text
  classification based on conditional reflection,'' \emph{IEEE Access}, vol.~7,
  pp. 76\,712--76\,719, 2019.

\bibitem{Ma_2015_ICCV}
L.~Ma, Z.~Lu, L.~Shang, and H.~Li, ``Multimodal convolutional neural networks
  for matching image and sentence,'' in \emph{The IEEE International Conference
  on Computer Vision (ICCV)}, December 2015.

\bibitem{CHEN2017221}
\BIBentryALTinterwordspacing
T.~Chen, R.~Xu, Y.~He, and X.~Wang, ``Improving sentiment analysis via sentence
  type classification using bilstm-crf and cnn,'' \emph{Expert Systems with
  Applications}, vol.~72, pp. 221 -- 230, 2017. [Online]. Available:
  \url{http://www.sciencedirect.com/science/article/pii/S0957417416305929}
\BIBentrySTDinterwordspacing

\bibitem{LiPruning}
H.~Li, A.~Kadav, I.~Durdanovic, H.~Samet, and H.~P. Graf, ``Pruning filters for
  efficient convnets.''

\bibitem{HeChannel}
Y.~He, X.~Zhang, and S.~Jian, ``Channel pruning for accelerating very deep
  neural networks.''

\bibitem{Liu2017Learning}
Z.~Liu, J.~Li, Z.~Shen, G.~Huang, S.~Yan, and C.~Zhang, ``Learning efficient
  convolutional networks through network slimming,'' pp. 2755--2763, 2017.

\bibitem{YeRethinking}
J.~Ye, X.~Lu, Z.~Lin, and J.~Z. Wang, ``Rethinking the
  smaller-norm-less-informative assumption in channel pruning of convolution
  layers.''

\bibitem{Chen2015Compressing}
W.~Chen, J.~T. Wilson, S.~Tyree, K.~Q. Weinberger, and Y.~Chen, ``Compressing
  neural networks with the hashing trick,'' \emph{Computer Science}, pp.
  2285--2294, 2015.

\bibitem{sironi2014learning}
A.~Sironi, B.~Tekin, R.~Rigamonti, V.~Lepetit, and P.~Fua, ``Learning separable
  filters,'' \emph{IEEE transactions on pattern analysis and machine
  intelligence}, vol.~37, no.~1, pp. 94--106, 2014.

\bibitem{sainath2013low}
T.~N. Sainath, B.~Kingsbury, V.~Sindhwani, E.~Arisoy, and B.~Ramabhadran,
  ``Low-rank matrix factorization for deep neural network training with
  high-dimensional output targets,'' in \emph{2013 IEEE international
  conference on acoustics, speech and signal processing}.\hskip 1em plus 0.5em
  minus 0.4em\relax IEEE, 2013, pp. 6655--6659.

\bibitem{jaderberg2014speeding}
M.~Jaderberg, A.~Vedaldi, and A.~Zisserman, ``Speeding up convolutional neural
  networks with low rank expansions,'' in \emph{Proceedings of the British
  Machine Vision Conference. BMVA Press}, 2014.

\bibitem{denil2013predicting}
M.~Denil, B.~Shakibi, L.~Dinh, M.~Ranzato, and N.~De~Freitas, ``Predicting
  parameters in deep learning,'' in \emph{Advances in neural information
  processing systems}, 2013, pp. 2148--2156.

\bibitem{hinton2015distilling}
G.~Hinton, O.~Vinyals, and J.~Dean, ``Distilling the knowledge in a neural
  network,'' \emph{arXiv preprint arXiv:1503.02531}, 2015.

\bibitem{luo2016face}
P.~Luo, Z.~Zhu, Z.~Liu, X.~Wang, and X.~Tang, ``Face model compression by
  distilling knowledge from neurons,'' in \emph{Thirtieth AAAI Conference on
  Artificial Intelligence}, 2016.

\bibitem{balan2015bayesian}
A.~K. Balan, V.~Rathod, K.~P. Murphy, and M.~Welling, ``Bayesian dark
  knowledge,'' in \emph{Advances in Neural Information Processing Systems},
  2015, pp. 3438--3446.

\bibitem{szegedy2016rethinking}
C.~Szegedy, V.~Vanhoucke, S.~Ioffe, J.~Shlens, and Z.~Wojna, ``Rethinking the
  inception architecture for computer vision,'' in \emph{Proceedings of the
  IEEE conference on computer vision and pattern recognition}, 2016, pp.
  2818--2826.

\bibitem{howard2017mobilenets}
A.~G. Howard, M.~Zhu, B.~Chen, D.~Kalenichenko, W.~Wang, T.~Weyand,
  M.~Andreetto, and H.~Adam, ``Mobilenets: Efficient convolutional neural
  networks for mobile vision applications,'' \emph{arXiv preprint
  arXiv:1704.04861}, 2017.

\bibitem{zhang2018shufflenet}
X.~Zhang, X.~Zhou, M.~Lin, and J.~Sun, ``Shufflenet: An extremely efficient
  convolutional neural network for mobile devices,'' in \emph{Proceedings of
  the IEEE Conference on Computer Vision and Pattern Recognition}, 2018, pp.
  6848--6856.

\bibitem{iandola2016squeezenet}
F.~N. Iandola, S.~Han, M.~W. Moskewicz, K.~Ashraf, W.~J. Dally, and K.~Keutzer,
  ``Squeezenet: Alexnet-level accuracy with 50x fewer parameters and< 0.5 mb
  model size,'' \emph{arXiv preprint arXiv:1602.07360}, 2016.

\bibitem{ma2018shufflenet}
N.~Ma, X.~Zhang, H.-T. Zheng, and J.~Sun, ``Shufflenet v2: Practical guidelines
  for efficient cnn architecture design,'' in \emph{Proceedings of the European
  Conference on Computer Vision (ECCV)}, 2018, pp. 116--131.

\bibitem{sandler2018mobilenetv2}
M.~Sandler, A.~Howard, M.~Zhu, A.~Zhmoginov, and L.-C. Chen, ``Mobilenetv2:
  Inverted residuals and linear bottlenecks,'' in \emph{Proceedings of the IEEE
  Conference on Computer Vision and Pattern Recognition}, 2018, pp. 4510--4520.

\bibitem{lin2013network}
M.~Lin, Q.~Chen, and S.~Yan, ``Network in network,'' \emph{arXiv preprint
  arXiv:1312.4400}, 2013.

\bibitem{Szegedy_2015_CVPR}
C.~Szegedy, W.~Liu, Y.~Jia, P.~Sermanet, S.~Reed, D.~Anguelov, D.~Erhan,
  V.~Vanhoucke, and A.~Rabinovich, ``Going deeper with convolutions,'' in
  \emph{The IEEE Conference on Computer Vision and Pattern Recognition (CVPR)},
  June 2015.

\bibitem{he2016deep}
K.~He, X.~Zhang, S.~Ren, and J.~Sun, ``Deep residual learning for image
  recognition,'' in \emph{Proceedings of the IEEE conference on computer vision
  and pattern recognition}, 2016, pp. 770--778.

\bibitem{goodfellow2013maxout}
I.~J. Goodfellow, D.~Warde-Farley, M.~Mirza, A.~Courville, and Y.~Bengio,
  ``Maxout networks,'' in \emph{Proceedings of the 30th International
  Conference on International Conference on Machine Learning-Volume 28}.\hskip
  1em plus 0.5em minus 0.4em\relax JMLR. org, 2013, pp. III--1319.

\bibitem{wu2018light}
X.~Wu, R.~He, Z.~Sun, and T.~Tan, ``A light cnn for deep face representation
  with noisy labels,'' \emph{IEEE Transactions on Information Forensics and
  Security}, vol.~13, no.~11, pp. 2884--2896, 2018.

\bibitem{molchanov2016pruning}
P.~Molchanov, S.~Tyree, T.~Karras, T.~Aila, and J.~Kautz, ``Pruning
  convolutional neural networks for resource efficient transfer learning,''
  \emph{arXiv preprint arXiv:1611.06440}, vol.~3, 2016.

\bibitem{simonyan2014very}
K.~Simonyan and A.~Zisserman, ``Very deep convolutional networks for
  large-scale image recognition,'' \emph{arXiv preprint arXiv:1409.1556}, 2014.

\bibitem{ioffe2015batch}
S.~Ioffe and C.~Szegedy, ``Batch normalization: Accelerating deep network
  training by reducing internal covariate shift,'' in \emph{International
  Conference on Machine Learning}, 2015, pp. 448--456.

\bibitem{glorot2011deep}
X.~Glorot, A.~Bordes, and Y.~Bengio, ``Deep sparse rectifier neural networks,''
  in \emph{Proceedings of the fourteenth international conference on artificial
  intelligence and statistics}, 2011, pp. 315--323.

\bibitem{mesaros2018multi}
A.~Mesaros, T.~Heittola, and T.~Virtanen, ``A multi-device dataset for urban
  acoustic scene classification,'' \emph{arXiv preprint arXiv:1807.09840},
  2018.

\bibitem{salamon2014dataset}
J.~Salamon, C.~Jacoby, and J.~P. Bello, ``A dataset and taxonomy for urban
  sound research,'' in \emph{Proceedings of the 22nd ACM international
  conference on Multimedia}.\hskip 1em plus 0.5em minus 0.4em\relax ACM, 2014,
  pp. 1041--1044.

\end{thebibliography}

%
%
%
%
%




\end{document}